\documentclass{article}

\usepackage{graphicx} 
\usepackage{booktabs}
\usepackage{amsmath}
\usepackage{subcaption}
\usepackage{array}
\usepackage{tabularx,ragged2e}
\usepackage{amssymb}
\usepackage{wrapfig}
\usepackage{xcolor}
\newcommand{\myparagraph}[1]{\par\noindent\textbf{#1}}

\usepackage[final]{corl_2026}

\title{GazeVLA: Learning Human Intention \\ for Robotic Manipulation}

\author{
    \normalfont Chengyang Li\textsuperscript{\normalfont 1,2,*} \quad
    Kaiyi Xiong\textsuperscript{\normalfont 1,*} \quad
    Yuan Xu\textsuperscript{\normalfont 3} \quad
    Lei Qian\textsuperscript{\normalfont 4} \quad
    Yizhou Wang\textsuperscript{\normalfont 3} \quad
    Wentao Zhu\textsuperscript{\normalfont 2} \\[0.5em]
    {\normalfont
    \textsuperscript{1} Shanghai Jiao Tong University \quad
    \textsuperscript{2} Eastern Institute of Technology, Ningbo} \\
    {\normalfont
    \textsuperscript{3} Peking University \quad
    \textsuperscript{4} ShanghaiTech University}
}

\hypersetup{
    pdftitle={Learning Human Intention for Robotic Manipulation},
    pdfauthor={Chengyang Li, Kaiyi Xiong, Yuan Xu, Lei Qian, Yizhou Wang, Wentao Zhu}
}

\begin{document}
\maketitle

\begin{figure}[h]
  \vspace{-10mm}
  \centering
  \includegraphics[width=1.0\linewidth]{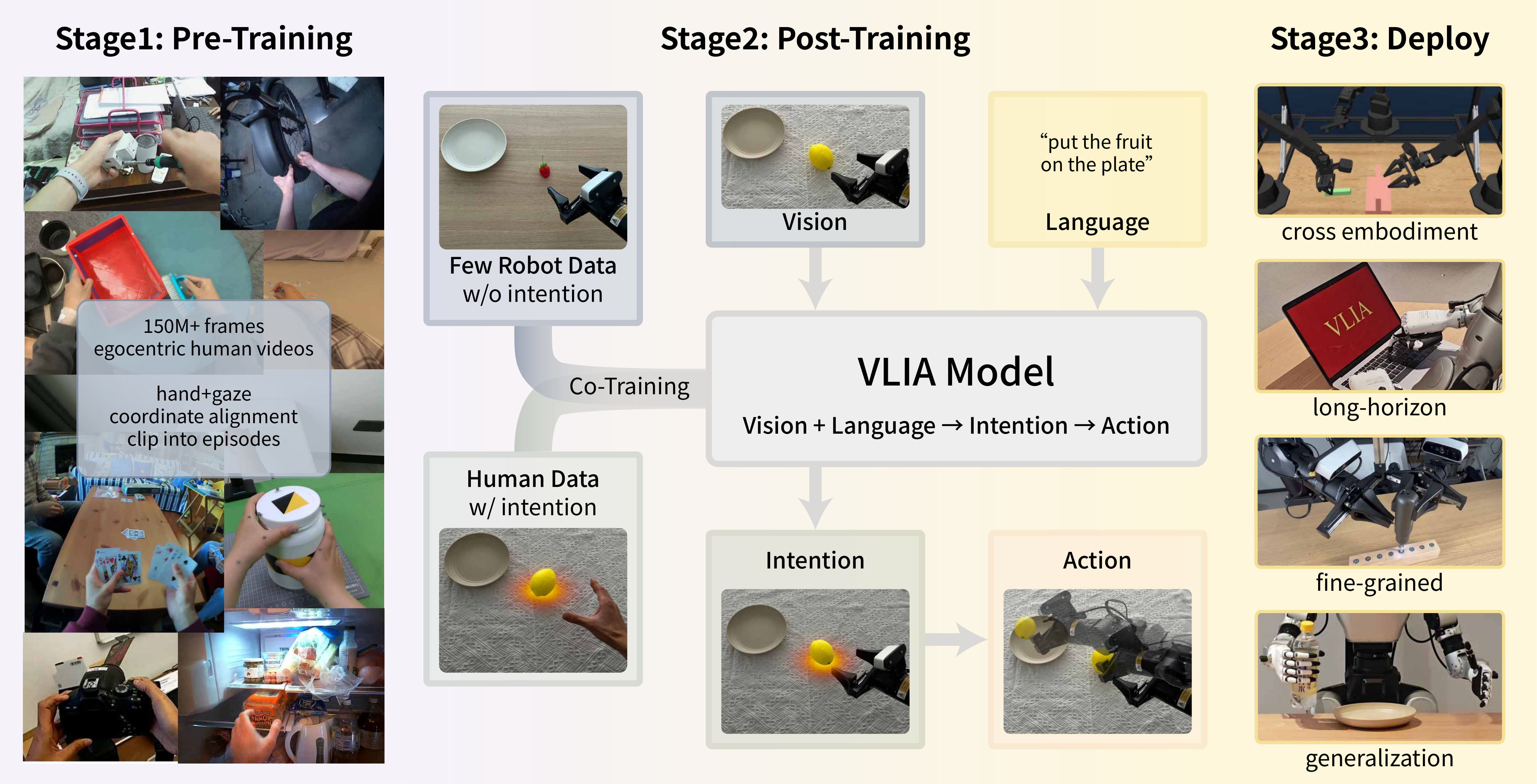}
  \caption{A human-to-robot learning framework that leverages human intention to facilitate robotic manipulation without requiring robot-side intention annotations. The model is first pre-trained on large-scale egocentric videos, followed by post-training on a limited set of robot and human demonstrations. We execute an intention-action reasoning chain during inference, enabling long-horizon, fine-grained tasks and robust generalization to unseen settings.}
  \label{fig:teaser}
  \vspace{-5mm}
\end{figure}

\begin{abstract}
Embodied foundation models have significantly propelled the advancement of artificial general intelligence, yet they rely heavily on large-scale data. Recent studies identify human behavior as one of the most scalable sources for fostering robotic physical intelligence. However, the inherent embodiment gap remains a fundamental challenge for effective human-to-robot learning.
In this work, we model human intention to enable efficient learning from human demonstrations. Among human behavioral cues, gaze simultaneously encodes high-level task semantics and exhibits a synergistic coupling with low-level action. By modeling intention through gaze, we derive a mid-level representation that facilitates cross-embodiment knowledge transfer.
We demonstrate that robots can internalize human intentions to enhance manipulation, even without robotic intention annotations. To capture the synergy between intention and action, we adopt a Chain-of-Thought reasoning paradigm during inference.
Our approach enables the execution of long-horizon and fine-grained tasks across diverse embodiments. Extensive experiments in both simulation and real-world demonstrate that our method consistently outperforms baseline approaches and generalizes to unseen settings. Project page: \href{https://gazevla.github.io/}{https://gazevla.github.io/}

\keywords{Learning from Human, Vision-Language-Action Model, Egocentric Human Videos, Robotic Manipulation}
\vspace{-4mm}

\end{abstract}

\section{Introduction}
\label{sec:intro}
\vspace{-4mm}

In recent years, embodied foundation models~\cite{openvla, pi0, pi05} have achieved remarkable progress toward general-purpose embodied intelligence. However, the success of these models heavily relies on large-scale robot data~\cite{oxe, bridgedatav2}, which is expensive to collect and difficult to scale, making data scarcity a fundamental bottleneck.

To alleviate this limitation, recent works~\cite{egomimic, egobridge, EMMA} have explored human data as a scalable and predictable supervision source for augmenting robot training data. Egocentric human videos~\cite{EgoDex, Ego4D, HOT3D} naturally include human behavioral intentions, task decomposition strategies, and object interaction priors, which are valuable for learning transferable manipulation skills. More importantly, compared with robot data, human data are substantially easier to collect, more diverse, and inherently scalable.

Early approaches~\cite{gu2023rttrajectory, A0, geminirobotics} leverage human data to learn explicit 2D visual representations, such as keypoints, affordances, and trajectories. However, these representations are typically obtained through post-processing and therefore fail to faithfully capture the underlying mechanisms that drive human behavior.
More recent methods~\cite{egovla, hrdt, egoscale, human2robot} jointly train human and robot as unified embodied agents, align action in latent spaces~\cite{LDA-1B, CLAP}, or employ world models for visual prediction~\cite{LAPA, UniVLA}. Nevertheless, these approaches primarily focus on direct action modeling, which is challenging due to the substantial embodiment gap between human and robot.

Motivated by insights from cognitive and behavioral studies~\cite{eye_hand_coordination, action_plans}, we argue that human behavior follows an intention-driven reasoning process. Compared with external action, internal intentions are more tightly coupled with the intrinsic motivations underlying behavior and are less sensitive to embodiment differences. We therefore introduce intention as an intermediate bridge between human and robot, enabling more effective human-to-robot knowledge transfer.
Furthermore, in natural human behavior, gaze not only attends to task-relevant regions but also temporally precedes action and exhibits strong correlations with subsequent behaviors. This property makes gaze a natural mid-level representation that jointly encodes high-level task semantics and low-level action details, making it an effective proxy for intention modeling.

Specifically, we first curate a large-scale egocentric human dataset for pre-training, enabling the model to learn the synergy between intention and action. During post-training, we use only a small amount of on-task human data together with robot data without intention annotations. During inference, the model follows a Chain-of-Thought reasoning process that first predicts intentions and then generates action, explicitly establishing the synergistic coordination between them.

We evaluate our approach on a reproducible simulation benchmark and six real-world robotic tasks. Experimental results demonstrate that our method enables both long-horizon and fine-grained manipulation across heterogeneous embodiments, outperforming the strongest baseline by 12.1\% in success rate. Furthermore, we show that intention modeling allows the robot to learn more efficiently from human data and generalize to unseen object placements, object categories, and scenes.

\vspace{-4mm}
\section{Related Work}
\label{sec:related_work}
\vspace{-4mm}

\myparagraph{Vision-Language-Action Models.}
Building upon the success of Vision-Language Models (VLM)~\cite{clip, llava, qwen, minigpt4}, Vision-Language-Action (VLA)~\cite{openvla,octo,pi0} models have made significant strides in embodied intelligence. 
Most existing VLA methods adopt either autoregressive tokenization strategies~\cite{openvla,rt2,rt1} or continuous action spaces modeled with flow-based generative paradigms~\cite{pi0,pi05,human2robot,Vlaser,GraspVLA,innon}. 
To further improve reasoning capabilities, recent studies have incorporated Chain-of-Thought (CoT) into models, either by leveraging task decomposition~\cite{pi05,ECoT,human2robot,DeepthinkVLA} or by predicting intermediate signals~\cite{ECoT,CoT-VLA,GraspVLA,acotvla} to enhance logical consistency.
However, the success of these approaches is largely attributed to the availability of extensive, high-quality robot data~\cite{oxe, bridgedatav2}, which is prohibitively expensive and difficult to scale.
In contrast, we treat humans as a distinct embodiment and leverage human data to scale robot training data.

\myparagraph{Mid-Level Representation.}
Recent studies~\cite{HiVLA} have demonstrated the effectiveness of leveraging mid-level conditioning representations to facilitate policy learning.  
Representative approaches~\cite{pi05, huang2022language} employ language-conditioned policies guided by task-level instructions. 
Other methods explore object-centric representations, including affordances~\cite{ardon2020affordances, rt-affordance}, keypoints~\cite{manuelli2019kpam}, and more action-centric representations, such as trajectories~\cite{gu2023rttrajectory} and motion flow~\cite{wen2023anypoint}.
Furthermore, recent methods~\cite{ye2026dreanzero, gao2026dreamdojo, kim2025cosmospolicy} leverage world models to predict target states for conditioning.
In contrast to prior approaches, we leverage human gaze as a transferable mid-level representation.
As a fundamental component of human behavior, gaze naturally attends to task-relevant regions and typically precedes physical actions. As a result, it is inherently correlated with high-level task semantics and low-level cues about subsequent actions.
Moreover, gaze signals can also be efficiently collected at scale using AR/VR devices~\cite{projectaria}.

\myparagraph{Learning from Human.}
To overcome the limitations of robot data scarcity, a growing body of work~\cite{egomimic, egobridge, EMMA, hommi, VITRA} has explored learning from egocentric human videos as a scalable supplement to robot demonstrations. Existing approaches can be broadly categorized into three paradigms.
The first line of work introduces explicit 2D visual primitives—such as keypoints~\cite{trackact, MimicPlay, masquerade, geminirobotics}, bounding boxes~\cite{geminirobotics}, or trajectories~\cite{wen2023anypoint, magma, gu2023rttrajectory}—as intermediate supervision signals. 
The second paradigm treats humans as an alternative embodiment, either by jointly training on human and robot data~\cite{human2robot, innon, egovla, hrdt, egoscale} or by aligning behaviors through a shared latent action space~\cite{CoMo, LAPA, UniVLA, villa-X, IGOR, ViPRA, Moto, GO-1, Grootn1, CLAP, motus, LAWN, LDA-1B}. 
A third line of research leverages human data to learn general visual representations~\cite{R3M, VIP, VC} or predictive world models~\cite{GR-1, GR-2, Gen2Act, mimic-video, FLARE, VPP} via large-scale visual pretraining. 
Compared to prior approaches, our work moves beyond learning what humans do to understanding why they do it. By explicitly modeling human intention instead of only imitating execution-level behaviors, we enable more effective human-to-robot knowledge transfer.
\begin{figure}[t]
  \centering
  \includegraphics[width=1.0\linewidth]{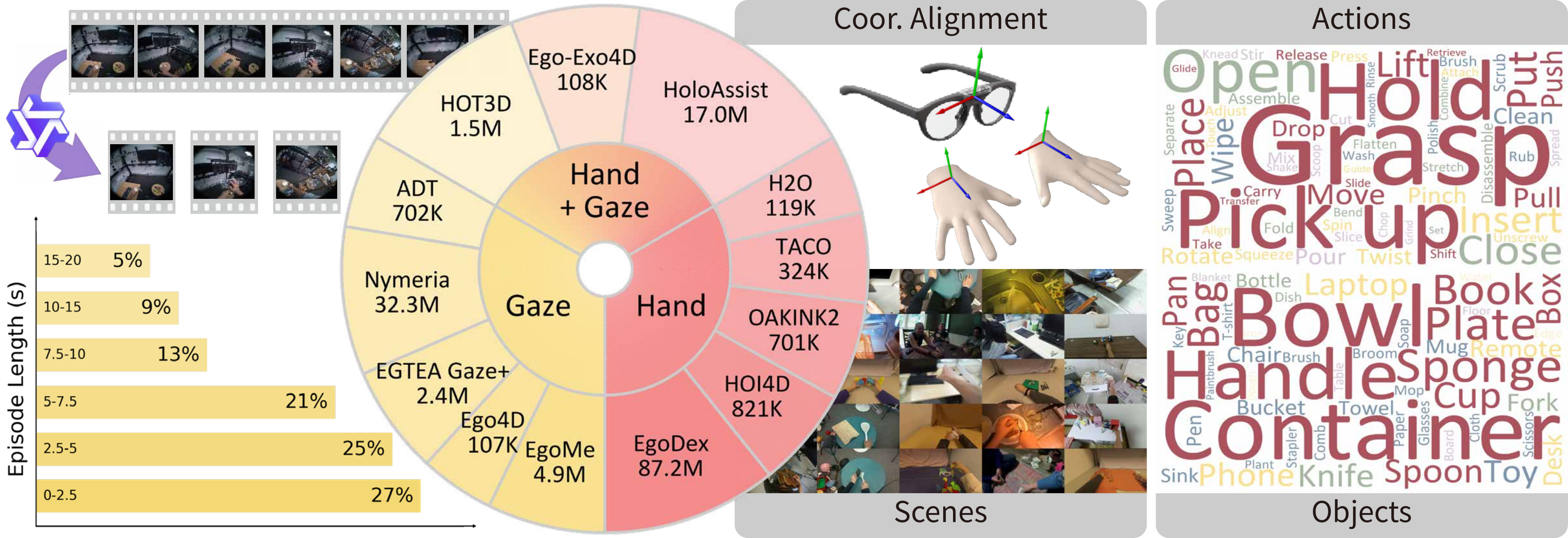}
  \caption{We curate a large-scale egocentric dataset aggregated from heterogeneous sources, containing hand and gaze annotations paired with validity masks. The dataset features a unified coordinate system and covers diverse scenes, actions, and objects. Long videos are clipped and filtered into short segments, resulting in over 150M high-quality motion frames.}
  \label{fig:dataset}
  \vspace{-5mm}
\end{figure}

\vspace{-4mm}
\section{Method} 
\label{sec:method}
\vspace{-4mm}

In this section, we present a human-to-robot learning framework that leverages human intention for robotic manipulation. 
We first introduce the data collection and processing pipeline in Sec.~\ref{sec:method_data}, then describe the model architecture in Sec.~\ref{sec:method_model}, and finally detail the training procedure in Sec.~\ref{sec:method_train}.

\vspace{-2mm}
\subsection{Data}
\label{sec:method_data}
\vspace{-2mm}

\myparagraph{Data Collection.}
We curate a large-scale egocentric dataset by aggregating 13 existing egocentric human datasets~\cite{EgoDex, Ego4D, HOT3D, ADT, Nymeria, EGTEAGaze+, EgoMe, Ego-Exo4D, HoloAssist, H2O, TACO, OAKINK2, HOI4D}. As shown in Fig.~\ref{fig:dataset}, the resulting dataset contains both hand and gaze annotations, covering more than 150M frames across diverse scenes, objects, and interaction types.
During post-training, we further collect on-task human demonstrations using AR glasses. Compared to robot data, human data collection is significantly more efficient and exhibits stronger generalization across objects, tasks, and scenes. By leveraging human data, our method enables robots to acquire capabilities that are never shown in robot data.

\myparagraph{Data Processing.}
We apply a unified preprocessing pipeline to standardize the heterogeneous datasets. Specifically, we employ Qwen~\cite{qwen} to segment long videos into shorter clips and discard redundant segments without hand-object interactions. We further refine the gaze annotations by discarding saccades and low-confidence samples. To ensure spatial consistency, we manually align the camera and wrist local coordinate systems, using the camera coordinate frame of the first frame in each clip as the reference frame.
We adopt the MANO~\cite{mano} model to represent hand motions. The resulting human action space $a \in \mathbb{R}^{48}$ consists of the positions of the five fingertip endpoints, together with the wrist position and rotation. Intention $i \in \mathbb{R}^{2}$ is modeled using gaze signals and represented as a single 2D pixel coordinate in the image plane. Binary masks are used to indicate the validity of hand and gaze annotations.
Additional details are provided in supplementary materials.

\begin{figure}[t]
  \centering
  \includegraphics[width=1.0\linewidth]{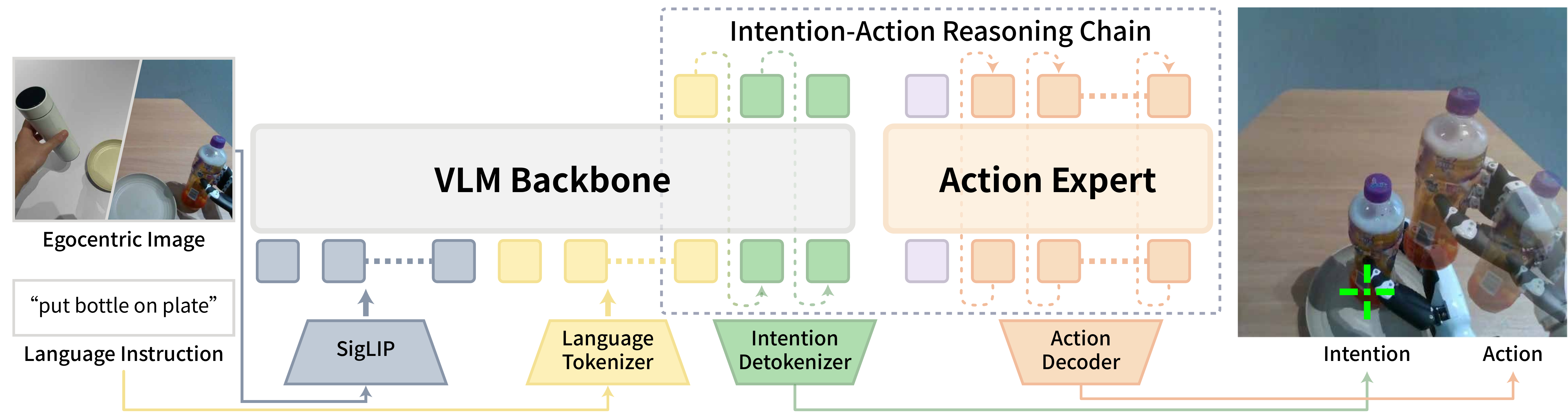}
  \caption{\textbf{Model Architecture}. VLIA model takes an egocentric image and language instructions as inputs, employing a Chain-of-Thought reasoning process to capture the synergy between intention and action. The model first leverages a VLM to predict gaze tokens in an autoregressive manner, followed by continuous action generation using flow matching. The intention is explicitly modeled via gaze as 2D image coordinates.}
  \label{fig:pipeline}
  \vspace{-5mm}
\end{figure}

\vspace{-2mm}
\subsection{VLIA Model}
\label{sec:method_model}
\vspace{-2mm}

\myparagraph{Model Architecture.}
The VLIA model takes a language instruction $l$, an egocentric camera image $o_{t}$ at time $t$, and the human hand or robot state $s_{t}$ as inputs. Conditioned on the visual observation and language instruction, the model first infers intention $i_{t}$ at time $t$ as an intermediate representation, and then predicts a sequence of future actions $a_{t:t+H}$ over a horizon of $H$ steps. We formalize this hierarchical prediction process as follows:
$$
\pi_{\theta} (a_{t:t+H}, i_{t} | s_{t}, o_{t}, l) = 
\pi_{\theta} (i_{t} | o_{t}, l) \cdot
\pi_{\theta} (a_{t:t+H} | i_{t}, s_{t}, o_{t}, l)
$$
An overview of our model architecture is presented in Fig.~\ref{fig:pipeline}. We utilize PaliGemma as our VLM backbone, which incorporates a SigLIP vision encoder and a Gemma-2B language model to process multimodal information. The action expert is designed to generate high-frequency, continuous actions through conditional flow matching.

\myparagraph{Intention-Action Reasoning Chain.}
Human behavior typically follows a structure in which intention precedes action. To explicitly capture this dependency, we introduce a Chain-of-Thought reasoning process that mirrors the human perception-intention-action paradigm.
We adopt gaze as an explicit proxy for intention, as it naturally precedes physical actions and provides strong cues for task-relevant visual regions and manipulation targets. To enable seamless integration with the VLM, we discretize 2D gaze coordinates into tokens through spatial binning.
As illustrated in Fig.~\ref{fig:pipeline}, during inference, the model first predicts intention tokens using next-token prediction within the VLM. Conditioned on the resulting KV cache, the action expert then generates continuous action sequences. 
Rather than directly imitating action, our framework introduces intention as a transferable mid-level representation that mitigates embodiment-specific action variations, thereby enabling more effective human-to-robot knowledge transfer.

\myparagraph{Loss Function.}
Our training objective includes an intention prediction loss for the VLM and an action generation loss for the action expert.
The intention loss is formulated as a standard autoregressive next-token prediction objective:
$$
\mathcal{L}_{\text{intent}} = 
\mathbb{E} \left[
-\sum_{n=1}^{N_{\text{intent}}} \log \pi_{\theta} (i^{n}_{t} | o_{t}, l, i^{<n}_{t})
\right]
$$
where $N_{\text{intent}}$ denote the length of the intention token sequences, and $i^{n}_{t}$ is the token at position $n$ in the sequences.
The action loss follows the conditional flow matching formulation:
$$
\mathcal{L}_{\text{action}} = 
\mathbb{E} \left[
|| 
\pi_{\theta}(a^{\tau}_{t:t+H}, \tau, c) - (a_{t:t+H} - \epsilon) 
||^{2}_{2}
\right]
$$
where 
$c = \{i_{t}, s_{t}, o_{t}, l\}$ represents the conditioning information,
$\tau \sim \mathcal{U}(0, 1)$ denotes flow matching timesteps, 
$\epsilon \sim \mathcal{N}(0, \mathbf{I})$ denotes Gaussian noise. 
The noisy action is constructed as $a^{\tau}_{t:t+H} = \tau a_{t:t+H} + (1-\tau) \epsilon$.
The final training objective is:
$$
\mathcal{L} 
= \lambda_{\text{action}}\mathcal{L}_{\text{action}} 
+ \lambda_{\text{intent}}\mathcal{L}_{\text{intent}}
$$

\vspace{-2mm}
\subsection{Training}
\label{sec:method_train}
\vspace{-2mm}

\myparagraph{Pre-Training.}
We pre-train the model on large-scale human data using a staged training strategy. During the warm-up stage, the vision encoder and VLM are frozen, followed by joint optimization of all parameters to stabilize training and prevent representation collapse.
To reduce the human-robot domain gap, we apply synchronized augmentations to both images and intention annotations, while downsampling human action sequences to match the robot control frequency.
Large-scale pre-training equips the model with strong priors for inferring intention from visual observations and task descriptions, while simultaneously capturing the synergy between intention and action.

\myparagraph{Post-Training.}
During the post-training stage, we jointly train the model on a small amount of human and robot data, where the robot data contain no intention annotations. We adopt a 1:1 sampling ratio between human and robot data. Robot actions are zero-padded and aligned with the length of the human actions. 
This post-training strategy enables the robot to internalize human intentions and acquire capabilities that are present only in human demonstrations.
More training details are provided in the supplementary materials.
\begin{figure}[t]
  \centering
  \includegraphics[width=1.0\linewidth]{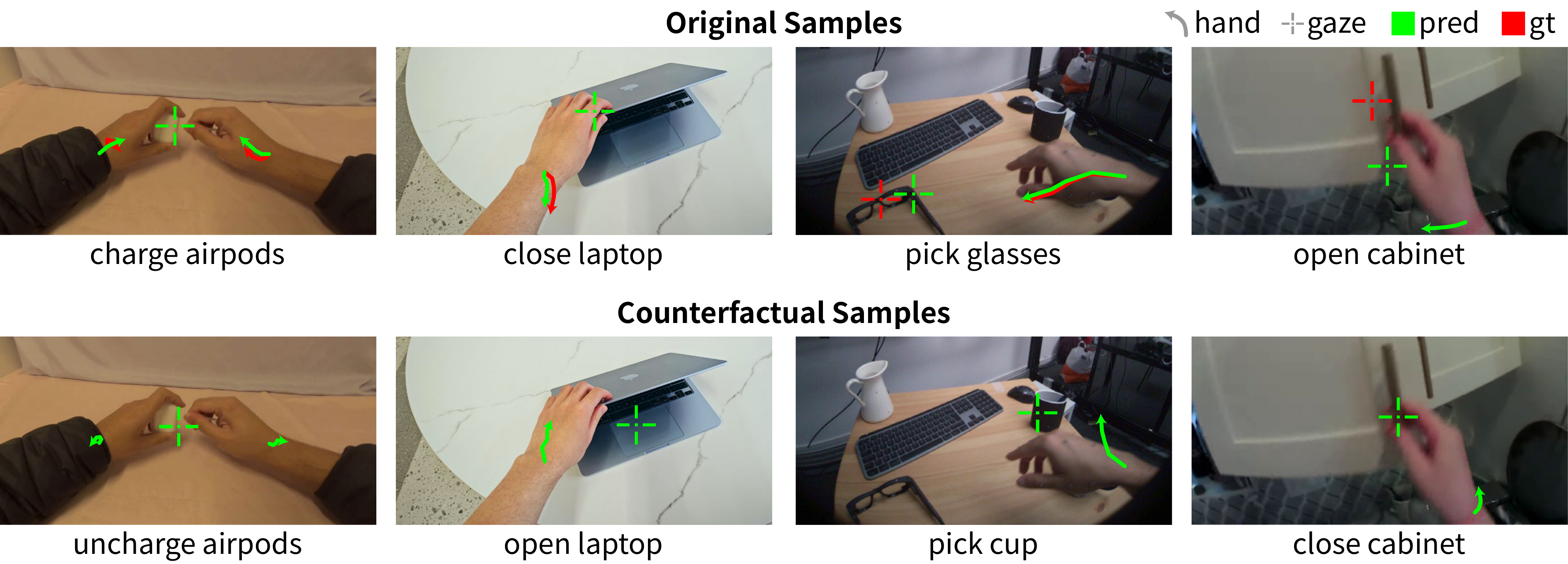}
  \caption{(Top): Original samples from the dataset. (Bottom): Counterfactual samples with modified instructions while keeping the same visual input. The results demonstrate the model’s ability to perform task-dependent reasoning.}
  \label{fig:pretrain_qualitative_analysis}
  \vspace{-5mm}
\end{figure}

\vspace{-4mm}
\section{Experiments}
\label{sec:exp}
\vspace{-4mm}

We conduct comprehensive evaluations of our approach from three perspectives:
(1) Can intention modeling improve robotic manipulation performance?
(2) Can human data facilitate robot learning and generalization?
(3) How do different intention sources affect policy performance?

\vspace{-2mm}
\subsection{Human Pre-training Evaluation}
\vspace{-2mm}
We first evaluate the model on human intention and hand motion prediction tasks. As shown in Fig.~\ref{fig:pretrain_qualitative_analysis}, the model generates plausible intention and action predictions conditioned on the task description. Quantitatively, the average intention prediction error is 4.8\% of the image diagonal, corresponding to 11 pixels for $224\times224$ images. For hand motion reconstruction, the model achieves a mean keypoint position error of 4.71 cm and a wrist rotation error of 12.31$^\circ$.

\vspace{-2mm}
\subsection{Simulation Experiments}
\vspace{-2mm}

\myparagraph{Benchmark and Baselines.}
We evaluate our method on the AV-ALOHA~\cite{av-aloha} benchmark, which integrates human gaze supervision with active visual perception. The platform comprises two 7-DoF arms for bimanual manipulation and a third 7-DoF camera arm for active vision.
We compare our approach against several representative baselines: 
LFA~\cite{lfa}, the official baseline provided by the AV-ALOHA~\cite{av-aloha} benchmark, which is built upon the ACT~\cite{act} architecture; 
HRDT~\cite{hrdt}, a diffusion-based human-to-robot learning framework; and
$\pi_{0.5}$~\cite{pi05}, a robot-pretrained VLA model.
For fair comparison, we follow the evaluation protocol of LFA~\cite{lfa}. Each model is trained using 100 trajectories per task and evaluated over 100 inference trials. To rigorously assess robustness, we additionally introduce visual distractors and lighting variations during evaluation, thereby testing the models' generalization capabilities under distribution shifts.

\begin{table}[t]
    \centering
    \small
    \resizebox{\linewidth}{!}{
    \begin{tabular}{lcccccccccccc}
        \toprule
        & \multicolumn{4}{c}{ID}
        & \multicolumn{4}{c}{OOD-Distractors}
        & \multicolumn{4}{c}{OOD-Lighting} \\
        
        \cmidrule(lr){2-5}
        \cmidrule(lr){6-9}
        \cmidrule(lr){10-13}

        Task $\backslash$ Method
        & LFA & HRDT & $\pi_{0.5}$ & \textbf{Ours}
        & LFA & HRDT & $\pi_{0.5}$ & \textbf{Ours}
        & LFA & HRDT & $\pi_{0.5}$ & \textbf{Ours} \\
        
        \midrule

        Cube Transfer
        & 87 & 89 & 94 & \textbf{100}
        & 12 & 28 & 25 & \textbf{35}
        & 0 & 12 & 32 & \textbf{36} \\

        Hook Package
        & 23 & 13 & 20 & \textbf{32}
        & 8 & 0 & 11 & \textbf{14}
        & 0 & 0 & 10 & \textbf{19} \\

        Peg Insertion
        & 10 & 17 & 15 & \textbf{18}
        & 7 & 2 & 6 & \textbf{9}
        & 0 & 0 & 7 & \textbf{16} \\

        Pour Test Tube
        & \textbf{41} & 24 & 34 & 39
        & 11 & 14 & 24 & \textbf{28}
        & 0 & 3 & \textbf{25} & 22 \\

        Slot Insertion
        & 43 & 42 & 54 & \textbf{60}
        & 23 & 19 & 47 & \textbf{56}
        & 0 & 11 & 44 & \textbf{50} \\

        Thread Needle
        & \textbf{56} & 46 & 33 & 43
        & 21 & \textbf{23} & 19 & \textbf{23}
        & 0 & 12 & 20 & \textbf{21} \\

        \midrule

        Average
        & 43 & 39 & 41 & \textbf{49}
        & 14 & 14 & 22 & \textbf{28}
        & 0 & 6 & 23 & \textbf{27} \\

        \bottomrule
    \end{tabular}
    }
    \captionsetup{skip=5pt}
    \caption{Quantitative comparison between our method and baseline methods including LFA~\cite{lfa}, HRDT~\cite{hrdt} and $\pi_{0.5}$~\cite{pi05} on the AV-ALOHA~\cite{av-aloha} benchmark. Our method achieves the best overall performance and demonstrates improved robustness across most out-of-distribution settings.}
    \label{tab:simulation_experiment}
    \vspace{-5mm}
\end{table}

\begin{figure}[t]
  \centering
  \includegraphics[width=1.0\linewidth]{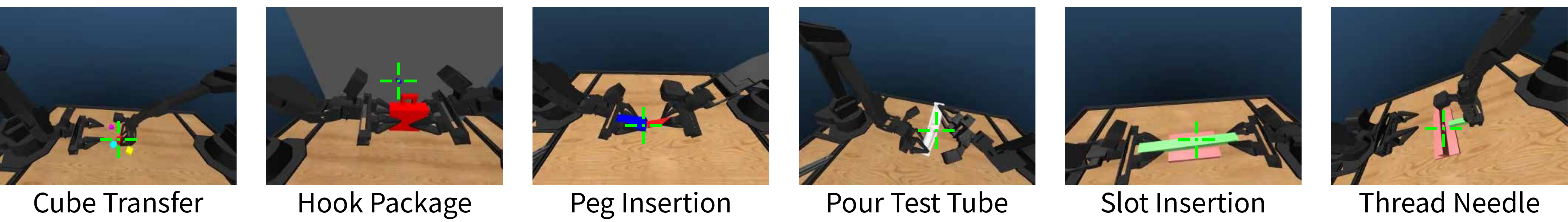}
  \caption{Qualitative evaluation of our method on the AV-ALOHA~\cite{av-aloha} benchmark. The green cross indicates the predicted intention point.}
  \label{fig:sim_experiment}
  \vspace{-5mm}
\end{figure}

\myparagraph{Comparison with Baselines.}
As shown in Tab.~\ref{tab:simulation_experiment}, our model outperforms all baselines under both in-distribution (ID) and out-of-distribution (OOD) settings. The improvement is particularly pronounced in the OOD scenario, where our method achieves a 22\% relative gain over $\pi_{0.5}$. 
Fig.~\ref{fig:sim_experiment} further demonstrates that intention prediction serves as an effective guidance signal for fine-grained manipulation tasks, including \textit{Hook Package} and \textit{Peg Insertion}, where precise spatial alignment is required.
Moreover, the intentions predicted by the VLM exhibit stronger robustness under distractor interference and lighting changes, resulting in improved generalization capability.

\vspace{-2mm}
\subsection{Real-World Experiments}
\vspace{-2mm}

\begin{figure}[t]
  \centering
  \includegraphics[width=1.0\linewidth]{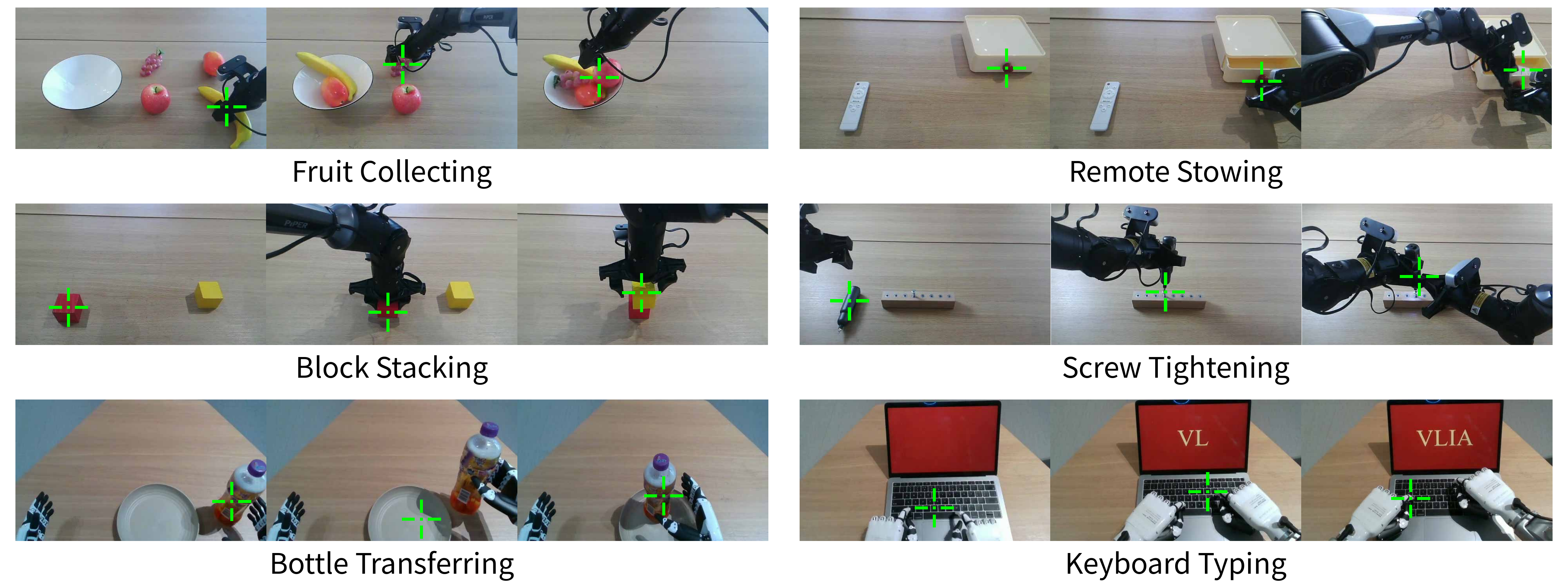}
  \captionsetup{skip=-1pt}
  \caption{Qualitative analysis of our method on real-world experiments. The green cross indicates the predicted intention.}
  \label{fig:real_task}
  \vspace{-5mm}
\end{figure}

\myparagraph{Real-World Robot and Task Setup.}
To evaluate cross-embodiment transfer and robustness in real-world robotic settings, we conduct extensive real-world experiments covering long-horizon tasks, fine-grained operations and, dexterous manipulation. Qualitative visualizations are presented in Fig.~\ref{fig:real_task}.
Specifically, the long-horizon tasks include \textit{Fruit Collecting} and \textit{Remote Stowing}. In \textit{Fruit Collecting}, the robot sequentially picks and places four fruits into a basket. \textit{Remote Stowing} requires the robot to open a drawer, place a remote controller inside, and close the drawer.
The fine-grained manipulation benchmark includes \textit{Block Stacking} and \textit{Screw Tightening}. \textit{Block Stacking} demands accurate alignment and placement of two blocks, while \textit{Screw Tightening} requires precise tool alignment and switch activation to complete the tightening process.
Dexterous manipulation tasks include \textit{Bottle Transferring} and \textit{Keyboard Typing}.
We evaluate our method on two real-world robotic platforms: a bimanual ALOHA robot with dual 7-DoF arms and grippers, and the Unitree G1 with dual 7-DoF arms and 6-DoF Inspire dexterous hands.
Additional details are provided in the supplementary materials.

\begin{figure}[t]
  \centering
  \includegraphics[width=1.0\linewidth]{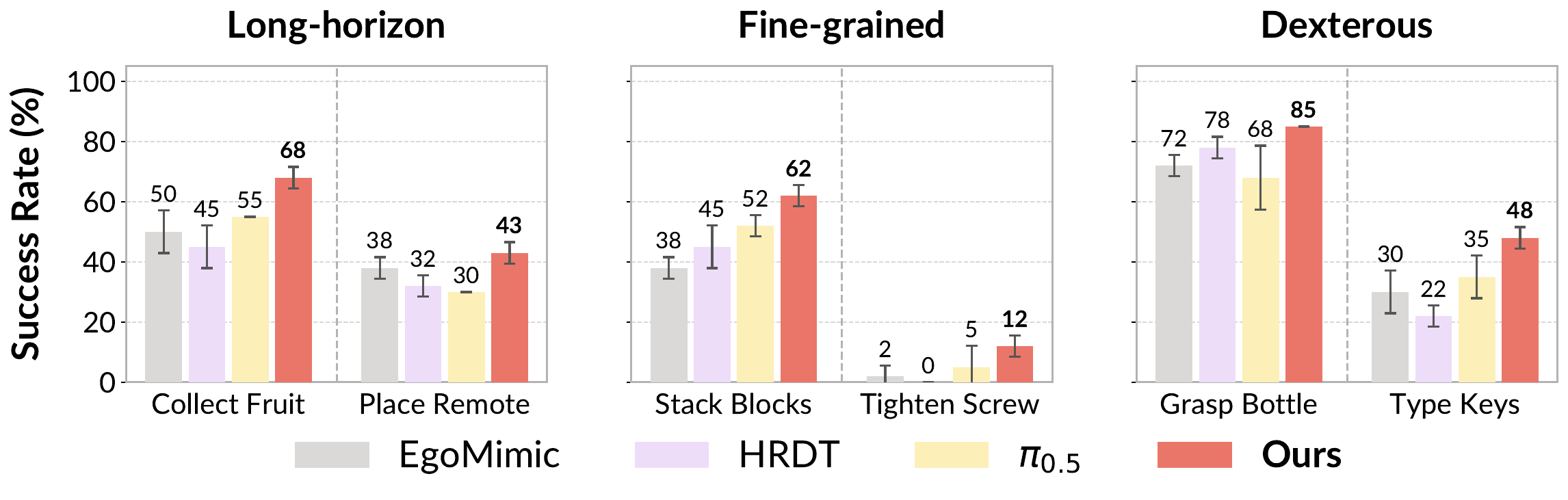}
  \captionsetup{skip=-1pt}
  \caption{Quantitative comparison between our method and baseline methods including EgoMimic~\cite{egomimic}, HRDT~\cite{hrdt} and $\pi_{0.5}$~\cite{pi05} on real-world experiments. Our method outperforms the baseline methods in long-horizon, fine-grained, and dexterous manipulation.}
  \label{fig:real_world_experiment}
  \vspace{-5mm}
\end{figure}

\myparagraph{Baselines and Results.}
We compare our method against three representative baselines: EgoMimic~\cite{egomimic}, a human-to-robot learning framework built upon the ACT~\cite{act} architecture; HRDT~\cite{hrdt}; and $\pi_{0.5}$~\cite{pi05}.
As shown in Fig.~\ref{fig:real_world_experiment}, our method consistently outperforms baseline approaches across long-horizon tasks, fine-grained manipulation, and dexterous manipulation, achieving an absolute success rate improvement of 12.1\% over $\pi_{0.5}$. 
In the long-horizon tasks \textit{Fruit Collecting} and \textit{Remote Stowing}, intention prediction effectively captures sub-task transitions across different interaction targets. Correspondingly, the generated actions adapt coherently with the evolving intentions, demonstrating stable coordination between intention and action.
In the fine-grained tasks \textit{Block Stacking} and \textit{Screw Tightening}, intention provides precise guidance for downstream action execution, enabling accurate pose alignment and tool interaction during manipulation.

\vspace{-2mm}
\subsection{Ablation Studies}
\vspace{-2mm}

\begin{table}[t]
    \centering
    \small
    \setlength{\tabcolsep}{10pt}
    \begin{tabular}{lcccc}
        \toprule
        Method & ID & OOD-Position & OOD-Object & OOD-Scene \\
        \midrule
        Ours 
        & \textbf{19/20} & \textbf{6/10} & \textbf{8/10} & \textbf{6/10} \\
        Ours w/o Intention 
        & 16/20 & 5/10 & 6/10 & 5/10 \\
        Ours w/o Human Pretrain 
        & 13/20 & 1/10 & 2/10 & 0/10 \\
        Ours w/o Human Finetune 
        & 17/20 & 3/10 & 5/10 & 2/10 \\
        \bottomrule
    \end{tabular}
    \captionsetup{skip=5pt}
    \caption{Ablation studies on pick-and-place task. We evaluate generalization across position shifts, object categories, and scene variations. The results demonstrate that intention modeling consistently improves manipulation performance, while human data play a significant role in both the pre-training and post-training stages.}
    \label{tab:ablation}
    \vspace{-5mm}
\end{table}

\begin{figure}[t]
    \centering
    \begin{minipage}{0.49\linewidth}
        \centering
        \includegraphics[width=\linewidth]{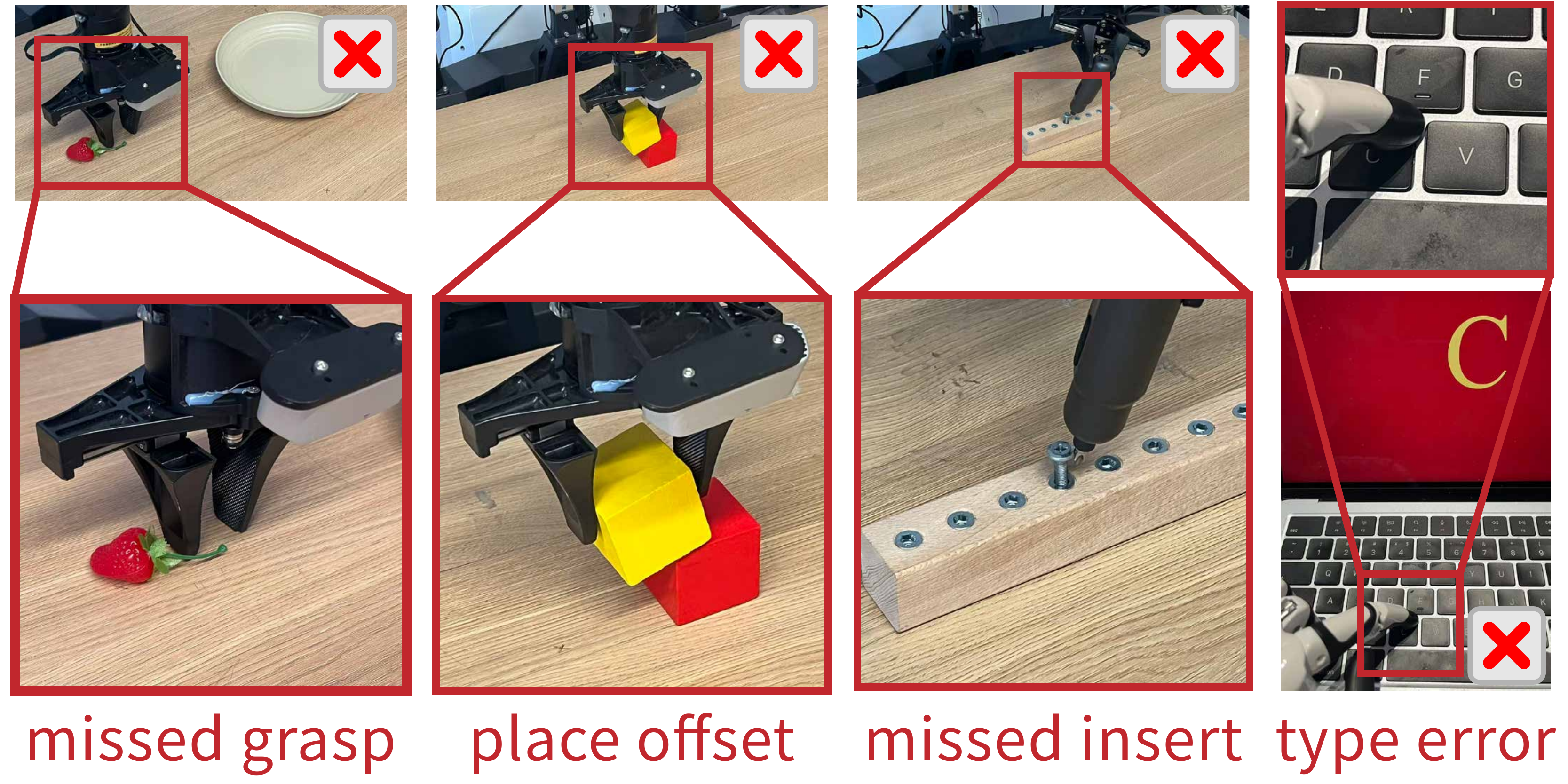}
        \caption{Representative failure cases of \textit{Ours w/o Intention}. Fine-grained manipulation tasks frequently exhibit action misalignment and execution failures without intention reasoning, highlighting the benefit of intention guidance.}
        \label{fig:failure_case}
    \end{minipage}
    \hfill
    \begin{minipage}{0.49\linewidth}
        \centering
        \includegraphics[width=\linewidth]{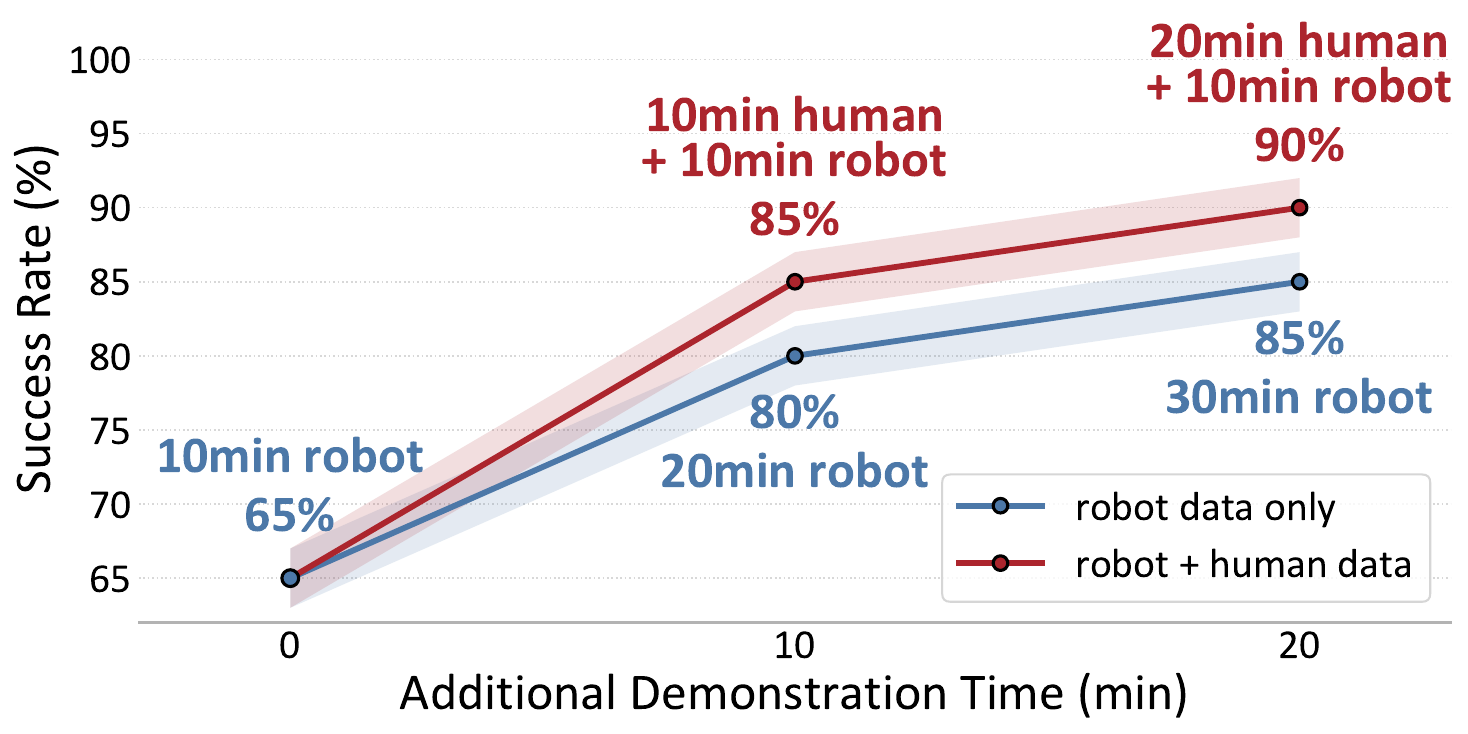}
        \caption{Comparison between human and robot data scaling on \textit{Bottle Transferring} task. With the same initial amount of robot data, adding human data yields substantially larger performance gains than adding robot data.}
        \label{fig:data_efficiency}
    \end{minipage}
    \vspace{-5mm}
\end{figure}

\myparagraph{Intention Enhances Manipulation and Generalization.}
We conduct ablation studies to validate the effectiveness of intention modeling. \textit{Ours w/o Intention} is trained on the same data as \textit{Ours}, but directly predicts actions without the intention-action reasoning chain.
Comparing \textit{Ours} and \textit{Ours w/o Intention} in Tab.~\ref{tab:ablation}, we observe that incorporating intention consistently improves task success rates in both in-domain and out-of-distribution settings.  
As shown in Fig.~\ref{fig:failure_case}, removing intention reasoning leads to noticeable misalignment in fine-grained manipulation tasks. This observation suggests that intention serves as a critical guidance signal for precise action execution. Moreover, intention also functions as an implicit task decomposition signal in long-horizon manipulation tasks and remains effective even when task descriptions are ambiguous, as in the \textit{Fruit Collecting} task.

\myparagraph{Human Data Enable Robot to Generalize to Unseen Positions, Objects, and Scenes.}
We evaluate the generalization capability of our method on the pick-and-place task. The robot data contain only a strawberry, with object placements restricted to a circular region of 5cm radius.
As shown in Tab.~\ref{tab:ablation}, introducing only human demonstration data collected under novel settings substantially improves the robot's generalization performance. In particular, the robot successfully operates within an expanded workspace of 8cm radius, manipulates previously unseen fruit categories such as lemons and cherries, and adapts to unfamiliar background environments. 
More details about the experimental setup can be found in the supplementary materials.

\myparagraph{Human Data Benefit Both Pre-training and Fine-tuning.}
We conduct ablation studies to analyze the role of human data.
\textit{Ours w/o Human Pretrain} removes human data during pre-training. In this setting, the model struggles to align intention prediction with action generation, indicating that large-scale human pre-training is crucial for learning the causal relationship between intention and action, as well as acquiring diverse visual priors.
\textit{Ours w/o Human Finetune} retains human pretraining but excludes human data during fine-tuning. We observe a clear degradation in generalization performance, suggesting that human demonstrations during fine-tuning are essential for improving robustness to unseen object positions, categories, and scenes.

\myparagraph{Human Data Scales Robot Performance More Efficiently.}
We conduct an efficiency comparison between human and robot data. As shown in Fig.~\ref{fig:data_efficiency}, we initialize training with 10 minutes of robot data and further augment it with either human or robot demonstrations collected under the same time budget.
We observe that human data yields substantially larger performance improvements. In particular, incorporating an additional 10 minutes of human demonstrations achieves gains comparable to those obtained from 20 additional minutes of robot data.
These results further highlight the scalability and data efficiency of our approach in leveraging human data.

\begin{wraptable}{r}{0.36\textwidth}
    \vspace{-2.5mm}
    \centering
    \captionsetup{skip=-0.2mm}
    \small
    \begin{tabular}{lc}
        \toprule
        Intention Source & Success Rate \\
        \midrule
        Predicted (Ours) & \textbf{65\%} \\
        Human Gaze       & 65\% \\
        VLM Keypoint    & 50\% \\
        No Intention     & 40\% \\
        Random Noise     & 25\% \\
        \bottomrule
    \end{tabular}
    \captionsetup{skip=5pt}
    \caption{Comparison of different intention sources.}
    \label{tab:intention_sources}
    \vspace{-4mm}
\end{wraptable}

\myparagraph{Effect of Different Intention Sources.}
We further study the effect of different intention sources on the \textit{Fruit Collecting} task. 
During inference, manually replacing the predicted intention leads to consistent changes in downstream action, indicating that our method actively conditions on intention.
As shown in Tab.~\ref{tab:intention_sources}, gaze substantially outperforms other intention sources, suggesting that it provides an anticipatory intention signal tightly coupled with downstream actions.
Please refer to the supplementary materials for more details.

\vspace{-4mm}
\section{Conclusion, Limitations and Future Directions}
\label{sec:conclusion}
\vspace{-4mm}

In this work, we facilitate human-to-robot learning by modeling human intentions. Specifically, intention is introduced as a transferable intermediate representation to mitigate embodiment discrepancies between human and robot, while the proposed intention-action reasoning chain explicitly captures the causal dependency.
Extensive experiments in both simulation and real-world settings demonstrate improved manipulation performance and generalization ability. Our results further suggest that multimodal signals in human behavior provide transferable supervision for robot learning.
Despite these promising results, several limitations remain. The chain-of-thought reasoning process introduces additional inference overhead, which may limit real-time deployment in highly dynamic scenarios. In addition, our method still relies on post-training data collection and does not yet achieve fully zero-shot transfer. Future work will explore more efficient reasoning mechanisms and more scalable learning from human videos toward zero-shot robotic manipulation.

\clearpage
\acknowledgments{This work was supported by the Zhejiang Provincial Natural Science Foundation of China under Grant No. LQN26F030037. The authors also thank the High Performance Computing Center at Eastern Institute of Technology, Ningbo, and the Ningbo Institute of Digital Twin for providing computational resources.
}
\bibliography{main}

\clearpage
\appendix
\section{Data}
\label{sup:data}

\subsection{Data Collection}
For on-task human data collection during post-training, we use Pupil glasses, which simultaneously capture RGB images, gaze signals, and camera intrinsic and extrinsic parameters. Human hand motion trajectories are obtained following the bimanual motion capture protocol of HaWoR, and are further processed with temporal smoothing and denoising filters to ensure trajectory continuity and data quality. We intentionally asked participants to perform robot-like actions and focus on task-relevant objects to reduce embodiment mismatch and make the data more useful.
During post-training, we collect on-task human demonstrations to improve robotic manipulation learning. To minimize the domain gap between humans and robots, we carefully align the task configurations and camera viewpoints with the robot setup, enabling more effective cross-embodiment transfer. On top of this aligned setting, we further vary object position, object categories, and environmental backgrounds in the human demonstrations. As a result, the learned policy can generalize to scenarios absent from the robot training data but observed in human demonstrations.

\subsection{Data Processing}
Most existing egocentric datasets consist of long-horizon videos without fine-grained action boundaries and often contain extensive redundant segments lacking hand interactions. To address this issue, we develop an automatic task-segmentation pipeline that decomposes long videos into short clips, each corresponding to a single atomic manipulation skill, while filtering out segments without hand-object interactions. 
Specifically, we adopt Qwen2.5-VL-7B-Instruct for video segmentation. Raw videos are uniformly sampled at 1 Hz, where each batch of 30 frames is resized to $224 \times 224$ before being processed by the VLM. We further design prompting templates to guide the model in identifying atomic action boundaries. The model outputs timestamp pairs corresponding to the start and end frames of each action segment, together with textual descriptions. All generated annotations are further standardized into a unified verb-noun format.

We observe that raw egocentric gaze data contain substantial noise, including frequent saccades and unstable gaze shifts, which can adversely affect model training. To improve gaze quality, we perform a dedicated gaze filtering procedure.
Concretely, we first employ heuristic criteria to identify valid fixation points, filtering out rapid gaze transitions, saccadic movements, and low-confidence gaze samples. Furthermore, neighboring gaze points with similar spatial locations are clustered into temporally stable fixation targets. Empirically, we find that this preprocessing strategy significantly improves gaze supervision quality and enhances the stability of intention prediction.

We employ masks to indicate the validity of action and intention labels. We use a two-dimensional binary mask to indicate the validity of the left and right hands, and a separate 1-dimensional mask to denote the validity of the gaze signal. During training, these masked elements are excluded from the attention mechanism and do not contribute to the loss calculation.

For human hand motion alignment and visualization, we first manually define a canonical coordinate system where the head camera in the first frame is placed at the origin, the human subject faces the positive X-axis, and the Z-axis points vertically upward. We then align the hand rotations such that the X-axis points from the wrist to the fingers and the Z-axis points from the wrist to the back of the hand. To accommodate datasets lacking MANO representations, we utilize hand keypoints as the network input. This choice allows us to incorporate more datasets and provides more accurate supervision than hand joint positions derived via inverse kinematics. In this work, all rotations are represented using the 6D rotation representation, as it is more amenable to stable model learning.

We represent the robot action space by jointly modeling the robot end-effector pose together with the gripper states or dexterous-hand joint angles. We use end-effector control rather than joint angles, which explicitly aligns the action spaces of the human wrist and the robot end-effector, enabling more effective knowledge transfer. 
We zero-pad the original robot action to match the 48-dimensional human motion representation. During training, we apply separate normalization for human and robot actions. Consistent with prior work, the model implicitly learns the correspondence between human and robot motions during training.

\section{Model}
\label{sup:model}

\begin{wrapfigure}{r}{0.4\textwidth}
    \centering
    \includegraphics[width=0.38\textwidth]{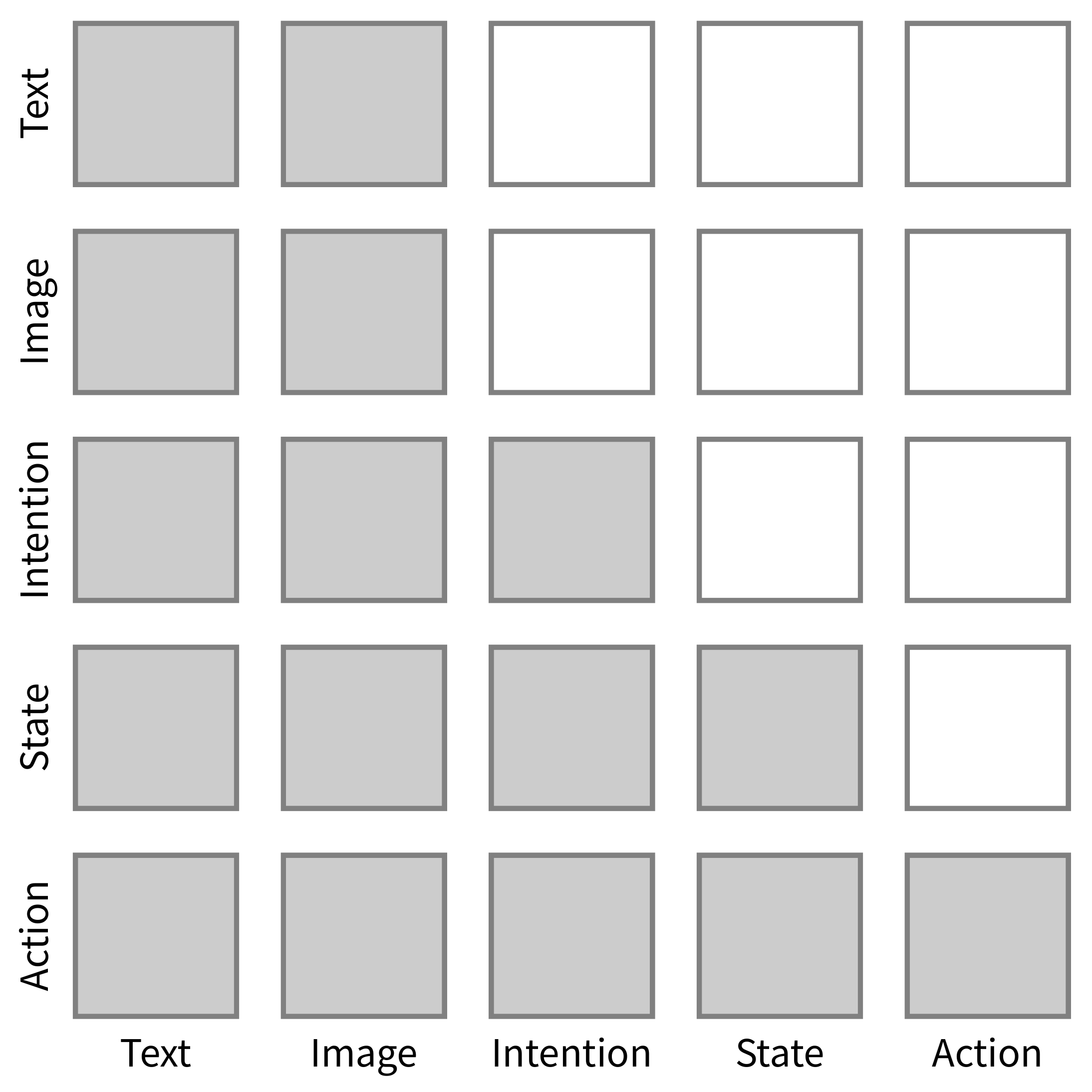} 
    \caption{Block-wise structured attention.}
    \label{fig:attention_mask}
    \vspace{-10mm}
\end{wrapfigure}

\subsection{Model Architecture}
Our model is built upon 3B PaliGemma-2 integrated with a flow-based action expert. The visual input consists of a single $224 \times 224$ image obtained via padding and resizing. During training, we apply consistent spatial augmentations (e.g., cropping and rotation) to both the visual inputs and their corresponding gaze coordinates. In addition, we introduce an encoder–decoder module appended to the intention feature to extract refined representations and predict the final intention probability.

Our model adopts a blockwise causal attention mask consisting of four token blocks: $[l, o_t], [i_t], [q_t], [a_t]$. Within each block, there is full bidirectional attention, whereas the tokens in each block cannot attend to the tokens in future blocks. The detailed attention structure is illustrated in Fig.~\ref{fig:attention_mask}.

\begin{table}[t]
\centering
\caption{Hyperparameters for pre-training and post-training.}
\label{tab:sup_training_details}
\begin{tabular}{l@{\hspace{8mm}}c@{\hspace{8mm}}c}
\toprule
Hyperparameter & Pre-Training & Post-Training \\
\midrule
Step & 20000 & 30000 \\
Batch Size & 2048 & 32 \\
Learning Rate & 5e-5 & 1e-4 \\
Learning Rate Scheduling & Constant & Cosine \\
Gradient Accumulation Steps & 4 & 1 \\
$\lambda_{\text{action}}$ & 1 & 1 \\
$\lambda_{\text{intent}}$ & 0.1 & 0.1 \\
\bottomrule
\end{tabular}
\end{table}

\subsection{Implementation Details}
As shown in Tab.~\ref{tab:sup_training_details}, we report the training configurations used for pre-training on large-scale egocentric human datasets and post-training on robot data.
During pretraining, our model is trained for 20k steps on 8 NVIDIA A800 GPUs, consuming a total of 1,344 GPU hours. We use a batch size of 2048 and set the learning rate to $5 \times 10^{-5}$. 
During pretraining, datasets are sampled proportionally to their sizes. In the post-training stage, we apply separate normalization for human and robot data and sample them at a balanced 1:1 ratio.
We employ a staged optimization strategy for pre-training. In the initial warm-up phase, we freeze both the vision encoder and the vision--language model, optimizing only the action expert module as well as the action encoder and decoder. Afterwards, all model parameters are jointly fine-tuned. Such a strategy stabilizes early-stage optimization, prevents representation collapse, and retains the pretrained capabilities of the foundation models.

We adopt a spatial binning strategy to discretize gaze coordinates into token representations. Given an input image of resolution 224×224, gaze locations are represented by pixel coordinates (h,w). The coordinates are first normalized into the continuous range [0,1]. We then uniformly divide both the horizontal and vertical dimensions into fixed spatial bins (\textit{num bins} = 224), quantizing continuous gaze coordinates into discrete spatial intervals. Each spatial bin is assigned a unique token ID based on its bin position, resulting in a discrete tokenized representation of gaze coordinates.
During inference, the predicted bin indices are mapped back to continuous spatial coordinates through inverse quantization. To support gaze prediction, we further use a dedicated linear decoder, which projects model hidden states into probability distributions over spatial bins for gaze coordinate estimation.

We adopt 2D keypoints to represent intentions instead of alternative formulations such as heatmaps or bounding boxes for two primary reasons. 
First, human gaze naturally manifests as sparse 2D fixation points, and our goal is to directly model this ground-truth behavioral signal.
Second, lightweight 2D point representations provide a more transferable abstraction for human-to-robot learning. In contrast to dense spatial representations such as heatmaps, 2D keypoints are easier to align across heterogeneous embodiments and can generalize more effectively under limited or even zero-shot robot supervision.

Due to natural head movements in egocentric human recordings, gaze points are heavily biased toward the image center. In contrast, attention targets in fixed-camera robotic setups are more uniformly distributed across the task workspace. To bridge this distribution gap, we apply synchronized augmentation to both visual observations and gaze annotations during training. Concretely, images and corresponding gaze points are jointly cropped, translated, and rotated, allowing the human gaze distribution to better align with the spatial distribution observed in robotic data.

\section{Simulation Experiments}
\label{sup:sim_exp}
\subsection{Training Configuration}
This section details the architectural adaptations and training configurations used for each baseline method. The AV-ALOHA benchmark contains six manipulation tasks, with each task providing 100 expert trajectories.
For fair evaluation, we establish a unified experimental protocol across all methods. We adopt a joint multi-task training setup, where a single policy is trained using demonstrations from all tasks and evaluated on each task individually at test time. All methods operate at the original control frequency of 25 FPS. The action chunk horizon is fixed to 25 steps, corresponding to predicting 1 second of future actions, and the predicted action sequence is executed in an open-loop manner after each inference step.
The action space is 21-dimensional, including dual-arm manipulation and active vision control. Detailed implementation and training settings for each baseline are provided below:
\begin{itemize}
    \item \textbf{LFA}: 100{,}000 training steps, batch size of 64, learning rate of $1\times10^{-4}$, Transformer hidden dimension of 512, 8 attention heads, 4 encoder layers, 7 decoder layers, prediction horizon of 25, dropout rate of 0.1, and action encoder/decoder dimension of 21.
    \item \textbf{HRDT}: 30{,}000 training steps, batch size of 16, learning rate of $1\times10^{-4}$, action encoder/decoder dimension of 21. All other hyperparameters are kept identical to the original implementation.
    \item \textbf{$\pi_{0.5}$}: 30{,}000 training steps, batch size of 32, and learning rate of $1\times10^{-4}$, action encoder/decoder dimension of 32 with zero padding.
\end{itemize}

\subsection{More Results}
\begin{figure}[t]
  \centering
  \includegraphics[width=1.0\linewidth]{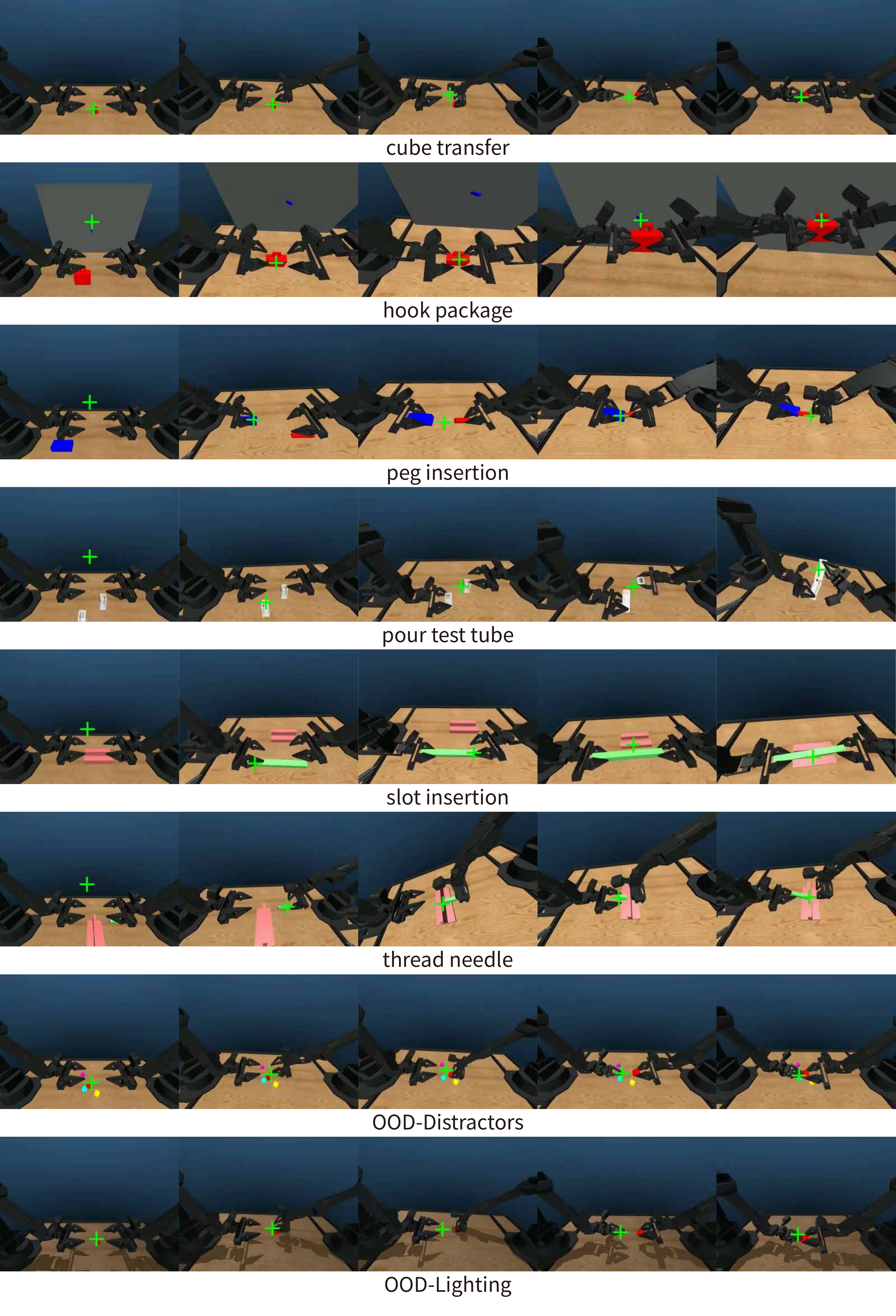}
  \caption{More qualitative evaluation of our method in simulation. The green cross indicates the predicted intention point.}
  \label{fig:sup_sim_exp}
\end{figure}

We further provide qualitative analysis on the AV-ALOHA benchmark. As shown in Fig.~\ref{fig:sup_sim_exp}, intention prediction serves as an effective guidance signal for downstream action execution. In the thread-needle task, the predicted intention consistently attends to the center of the hole, enabling successful insertion through the narrow opening. Similarly, in the hook-package task, our method accurately guides the backpack toward the hook for precise placement.
We also observe that VLM-predicted intentions are considerably more robust under environmental perturbations, including distractor objects and lighting variations, which further improves the overall generalization capability of the policy. 

\clearpage

\section{Real-World Experiments}
\label{sup:real_exp}
\subsection{Task Description}
Tab.~\ref{tab:real_world_task_description} provides detailed information about the real-world experiments, including the task procedures, success criteria, and training data collection protocols for each task. For each method and task, we conduct two evaluation rounds with 20 trials per round, totaling 2 rounds of 20 trials (40 total). We report the mean and standard deviation of the success rates across the two rounds.

\begin{table}[htbp]
\centering
\caption{Real-World Task Descriptions and Evaluation Protocols}
\label{tab:real_world_task_description}
\renewcommand{\arraystretch}{1.2}
\begin{tabular}{|p{3cm}|p{4.7cm}|p{4.7cm}|}
\hline
\textbf{Task Name} & \textbf{Task Description} & \textbf{Evaluation Protocols} \\
\hline
Fruit Collecting &
Pick up all fruits on the table sequentially and place them into the plate. &
100 robot data; 200 human data; 2 rounds of 20 trials (40 total); success if all fruits are placed in the bowl. \\
\hline

Remote Stowing &
Open the drawer, pick up the remote control, place it into the drawer, and close the drawer. &
50 robot data; 100 human data; 2 rounds of 20 trials (40 total); success if the remote control is inside the drawer. \\
\hline

Block Stacking &
First place the red block in the center, then stack the yellow block on top of it. &
50 robot data; 100 human data; 2 rounds of 20 trials (40 total); success if the yellow block is on top of the red block. \\
\hline

Screw Tightening &
The left arm picks up the screwdriver and aligns it with the screw, while the right arm presses the electric screwdriver switch. &
100 robot data; 200 human data; 2 rounds of 20 trials (40 total); success if the screw is tightened into the wooden block. \\
\hline

Bottle Transferring &
Use the dexterous hand to pick up the bottle and place it onto the plate. &
20 robot data; 100 human data; 2 rounds of 20 trials (40 total); success if the bottle is stably placed on the plate. \\
\hline

Keyboard Typing &
Press the four letters ``V'', ``L'', ``I'', and ``A'' sequentially on the keyboard. &
20 robot data; 100 human data; 2 rounds of 20 trials (40 total); success if the screen displays ``VLIA''. \\
\hline
\end{tabular}
\end{table}

\subsection{Training Configuration}
We detail the training setups for EgoMimic, HRDT, $\pi_{0.5}$, and our approach. For training data, EgoMimic and our method are jointly optimized using both human and robot data, while HRDT and $\pi_{0.5}$ are trained exclusively on robot demonstrations. This design choice follows the original implementations of HRDT and $\pi_{0.5}$, which do not support joint human--robot post-training.
As $\pi_{0.5}$ is pretrained with a predefined robot action space, we adopt a joint-angle action representation to preserve compatibility with its original architecture. In contrast, EgoMimic, HRDT, and our method all employ the unified action representation introduced in this work, with corresponding modifications to the action encoder and decoder dimensions.

\subsection{More Results}
\begin{figure}[t]
  \centering
  \includegraphics[width=1.0\linewidth]{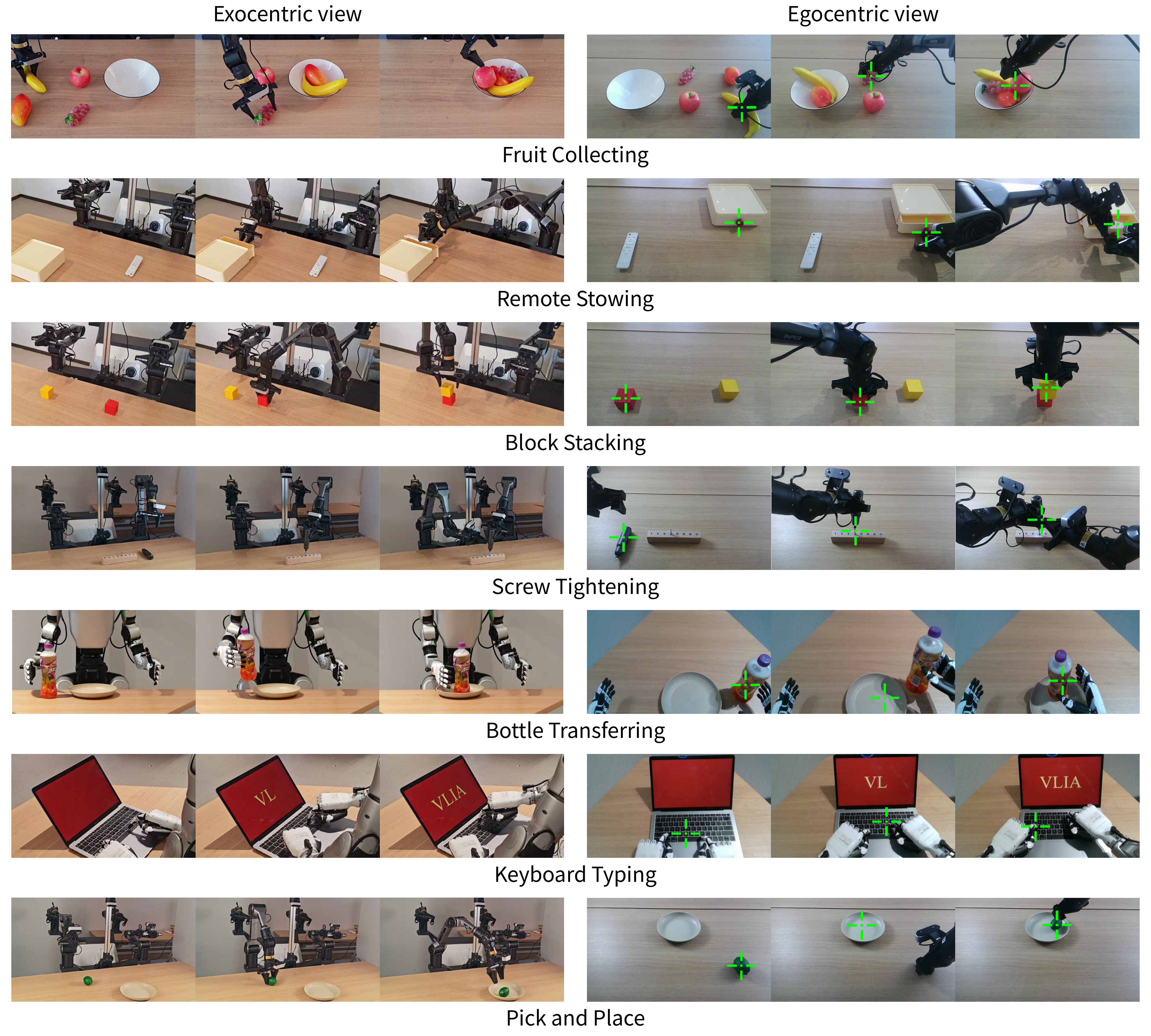}
  \caption{More qualitative evaluation of our method in real-world setting. The green cross indicates the predicted intention point. The left column shows the third-person view, while the right column presents the egocentric first-person view.}
  \label{fig:sup_real_exp_demo}
\end{figure}

We further present additional visualizations from real-world experiments. As shown in Fig.~\ref{fig:sup_real_exp_demo} the left side corresponds to the third-person perspective, while the right side shows the egocentric first-person observations. More qualitative results are available on the project website and in the accompanying video.

\clearpage

\begin{figure}[t]
  \centering
  \includegraphics[width=1.0\linewidth]{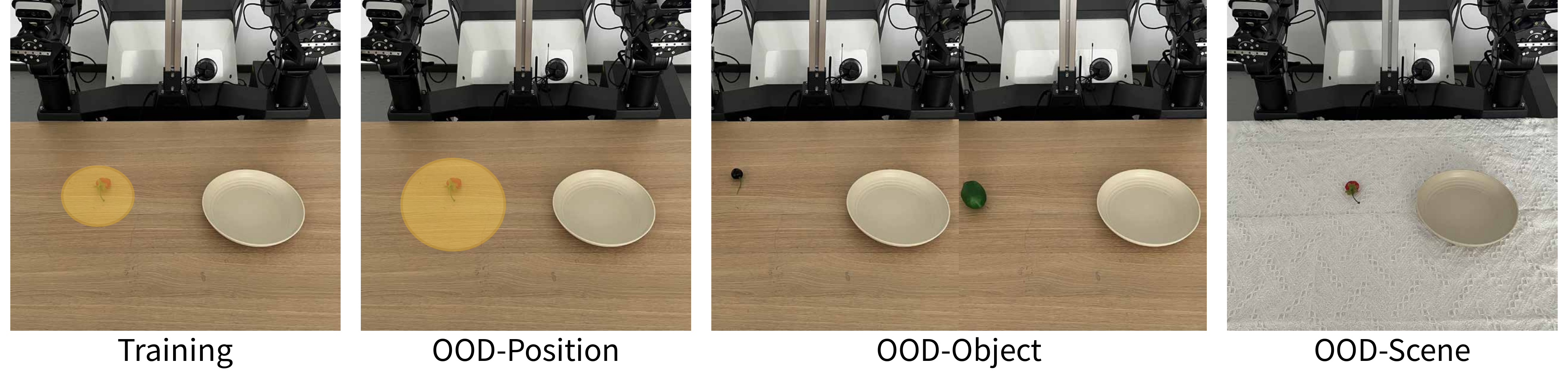}
  \caption{Generalization evaluation setup on pick-and-place task. The orange circle indicates the initial object positions.}
  \label{fig:pick_and_place_setting}
\end{figure}

\section{Results}
\label{sup:results}

\subsection{Human Data Enable Robot to Generalize to Unseen Positions, Objects, and Scenes}
\label{sup:ood_setting}
We further detail the OOD settings used in the ablation experiments for the pick-and-place task. As shown in Fig.~\ref{fig:pick_and_place_setting}, the robot training set contains only strawberry-grasping demonstrations, where object placements are constrained within a 5cm radius region. We evaluate OOD generalization along three dimensions: position, object category, and scene background. 
Specifically, position generalization tests objects randomly placed within a larger 8cm radius region; object generalization includes unseen fruits such as lemons and cherries; and scene generalization introduces a white tablecloth as a novel background perturbation.

\begin{minipage}[t]{0.48\textwidth}
    \centering
    \includegraphics[width=\linewidth]{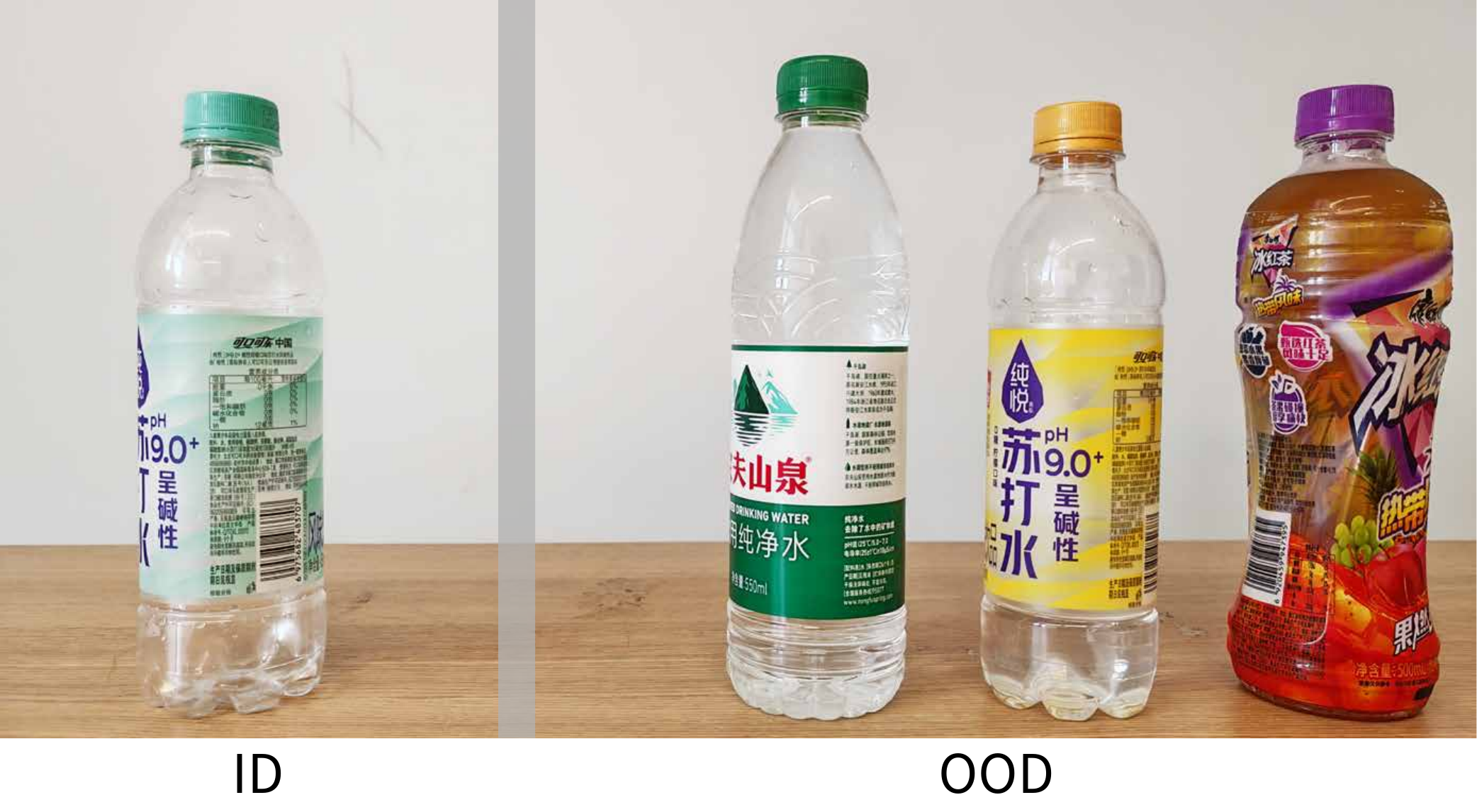}
    \captionof{figure}{Evaluation protocols on \textit{Bottle Transferring} task.}
    \label{fig:bottle_transfer_setting}
\end{minipage}
\hfill
\begin{minipage}[t]{0.48\textwidth}
    \vspace{0pt}

    \centering
    \small
    \vspace{-100pt}

    \captionof{table}{Comparison between human and robot data scaling on \textit{Bottle Transferring} task.}
    \label{tab:human_robot_efficiency}
    \begin{tabular}{lcc}
        \toprule
        Data & ID & OOD \\
        \midrule
        10m robot & 4/5 & 9/15 \\
        20m robot & 4/5 & 12/15 \\
        10m human + 10m robot & 5/5 & 12/15 \\
        30m robot & 5/5 & 12/15 \\
        20m human + 10m robot & 5/5 & 13/15 \\
        \bottomrule
    \end{tabular}
\end{minipage}

\subsection{Human Data Scales Robot Performance More Efficiently}
This section details the experimental setup for evaluating the relative data efficiency of human and robot demonstrations. Experiments are conducted on the dexterous manipulation task \textit{Bottle Transferring}. As shown in Fig.~\ref{fig:bottle_transfer_setting}, the blue bottle on the left is shared between both human and robot training data, while bottles with unseen colors and geometries on the right appear exclusively in human demonstrations and are never observed in robot data, forming the OOD evaluation setting.
The quantitative comparison is summarized in Tab.~\ref{tab:human_robot_efficiency}. Results show that human demonstrations provide substantially higher data efficiency than robot teleoperation data. Under the same 10-minute collection budget, robot teleoperation acquires only around 20 trajectories, whereas human demonstrations can provide approximately 100 trajectories.

\subsection{Effect of Different Intention Sources}
We additionally study the impact of different intention sources during inference. The configurations are defined as follows.

\begin{itemize}
\item \textbf{Predicted (Ours)}: intention tokens are generated by the proposed autoregressive intention predictor.

\item \textbf{Human Gaze}: human gaze signals collected using AR glasses synchronized with the robot viewpoint are directly used as intention inputs.

\item \textbf{VLM Keypoint}: a vision--language model predicts task-relevant 2D keypoints from image observations and language instructions, which are then used as intention signals.

\item \textbf{No Intention}: the intention pathway is disabled during inference by masking the intention-related attention mechanism, resulting in direct action prediction without intention reasoning.

\item \textbf{Random Noise}: randomly sampled spatial points are injected as noisy intention signals.
\end{itemize}

As shown in the results, both predicted intentions and human gaze consistently achieve the best performance, indicating that these signals better capture the intrinsic structure of human intentions. While VLM-predicted keypoints can accurately localize target objects, they do not possess the anticipatory temporal characteristics of human gaze, leading to slightly weaker performance. Removing intention reasoning causes unstable manipulation behaviors, especially during grasping and placement, while random noisy intentions further degrade performance. These observations collectively suggest that the learned policy meaningfully depends on intention guidance for downstream action generation.

\begin{figure}[t]
  \centering
  \includegraphics[width=1.0\linewidth]{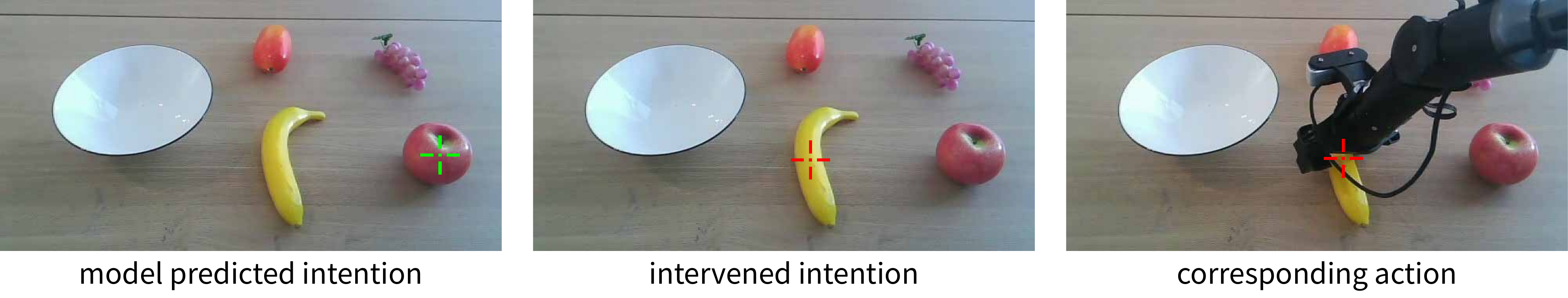}
  \caption{We evaluated the model's adherence to intention. During reasoning in the \textit{Fruit Collecting} task, we manually modified the predicted intention, and the results showed that the actions followed the guidance of the intention and grasped the corresponding objects. The green crosses denote the model predicted intentions, while the red crosses indicate the intervened intentions.}
  \label{fig:intention_guidance}
\end{figure}

\subsection{Synergy Between Intention and Action}
We find that intentions exhibit strong coordination with downstream actions. During inference on the \textit{Fruit Collecting} task, we manually modified the predicted intentions and, as shown in Fig.~\ref{fig:intention_guidance}, observed that the generated actions consistently followed the intention guidance to grasp the corresponding target objects. This result indicates that the policy actively relies on intention signals during action generation.

For long-horizon tasks such as \textit{Keyboard Typing}, intention acts as an implicit task decomposition signal that guides sequential action execution over target keys. In the \textit{Remote Stowing} task, the predicted intention dynamically transitions across different interaction targets throughout the manipulation sequence.
We observe similar effects in fine-grained manipulation tasks such as \textit{Screw Tightening}, where accurate intention localization significantly improves precise alignment and leads to substantially higher success rates.

\subsection{Tradeoff between Sensitivity and Robustness on Intention Guidance}
We identify an inherent trade-off between \textit{sensitivity} and \textit{robustness}. Sensitivity characterizes how strongly the policy attends to intention signals, leading to action generation that closely follows the predicted intentions. Robustness, on the other hand, reflects the policy's ability to recover from inaccurate intention predictions and still complete the task successfully.
For tasks with underspecified language instructions, such as \textit{Collect Fruits}, intention prediction provides informative guidance for selecting manipulation targets. However, conflicts between language instructions and intention signals expose the trade-off between sensitivity and robustness. For instance, the instruction may specify grasping the left fruit while the predicted intention attends to the right fruit, or the intention input may be replaced entirely with noise. Under these conditions, task performance drops significantly, suggesting that the policy meaningfully depends on intention information for downstream decision-making.
We further find that this balance between sensitivity and robustness can be adjusted through the weighting of the training losses.

\subsection{Inference Latency}
We measure inference latency on an Intel Core i9-14900KF CPU and an NVIDIA RTX 4090 GPU in the 10\,Hz control setting. Our method takes 44.384\,ms per inference, compared with 32.205\,ms for direct action prediction, corresponding to an additional 12.179\,ms. 

\end{document}